\documentclass{article}
\usepackage{colortbl}
\usepackage{array}
\usepackage{multirow}
\usepackage{makecell}

\usepackage{caption}
\usepackage{wrapfig}

\newcommand{\incre}[1]{\textcolor{teal!90}{#1}}
\usepackage{microtype}
\usepackage{subcaption}
\usepackage{graphicx}
\usepackage{booktabs} 

\usepackage{hyperref}

\usepackage[ruled,vlined,linesnumbered]{algorithm2e}

\usepackage[accepted]{icml2025}

\usepackage{amsmath}
\usepackage{amssymb}
\usepackage{mathtools}
\usepackage{amsthm}

\usepackage[capitalize,noabbrev]{cleveref}

\theoremstyle{plain}

\theoremstyle{definition}

\theoremstyle{remark}

\usepackage[textsize=tiny]{todonotes}

\icmltitlerunning{Compositional Attribute Imbalance in Vision Datasets}

\begin{document}

\twocolumn[
\icmltitle{Compositional Attribute Imbalance in Vision Datasets}




\icmlsetsymbol{equal}{*}

\begin{icmlauthorlist}
\icmlauthor{Jiayi Chen}{equal,yyy}
\icmlauthor{Yanbiao Ma}{equal,sch}
\icmlauthor{Andi Zhang}{comp}
\icmlauthor{Weidong Tang}{yyy}
\icmlauthor{Wei Dai}{yyy}
\icmlauthor{Bowei Liu}{thu}


\end{icmlauthorlist}

\icmlaffiliation{yyy}{Xidian University}
\icmlaffiliation{comp}{Centre for AI Foundamentals, University of Manchester}
\icmlaffiliation{sch}{Renmin University of China}
\icmlaffiliation{thu}{Tsinghua University}

\icmlcorrespondingauthor{Yanbiao Ma}{ybma1998xidian@gmail.com}
\icmlcorrespondingauthor{Andi Zhang}{az381@cantab.ac.uk}

\icmlkeywords{Machine Learning, ICML}

\vskip 0.3in
]



\printAffiliationsAndNotice{\icmlEqualContribution} 

\begin{abstract}
Visual attribute imbalance is a common yet underexplored issue in image classification, significantly impacting model performance and generalization. In this work, we first define the first-level and second-level attributes of images and then introduce a CLIP-based framework to construct a visual attribute dictionary, enabling automatic evaluation of image attributes. By systematically analyzing both single-attribute imbalance and compositional attribute imbalance, we reveal how the rarity of attributes affects model performance. To tackle these challenges, we propose adjusting the sampling probability of samples based on the rarity of their compositional attributes. This strategy is further integrated with various data augmentation techniques (such as CutMix, Fmix, and SaliencyMix) to enhance the model's ability to represent rare attributes. Extensive experiments on benchmark datasets demonstrate that our method effectively mitigates attribute imbalance, thereby improving the robustness and fairness of deep neural networks. Our research highlights the importance of modeling visual attribute distributions and provides a scalable solution for long-tail image classification tasks.
\end{abstract}

\section{Introduction}
\label{Introduction}

Addressing data imbalance in computer vision tasks remains a core challenge for improving model performance \cite{paper58,ma2025predicting}. Imbalances in the number of training samples across classes often lead to biases during the learning process, making it difficult for deep learning models to accurately recognize underrepresented classes. To tackle this issue, researchers have proposed various approaches, such as class-aware sampling strategies, loss reweighting, and balanced data augmentation techniques \cite{paper37,paper45,paper36,paperma,paperma1,paper38,paper39,paper75,paper76,paper46}. However, these methods primarily focus on inter-class imbalances, assuming that achieving balance at the class level suffices to ensure fairness and efficacy in learning. This assumption, however, overlooks intra-class attribute imbalances, particularly the problem of compositional attribute imbalance.

Attribute imbalance refers to the uneven distribution of image attributes (e.g., color, texture, and shape) within a single class. This imbalance can bias the learned representations of a model. While limited studies \cite{tang2022invariant,liu2021handling} have qualitatively discussed the challenges posed by attribute imbalance, no prior research has systematically quantified or analyzed its prevalence, severity, and impact on model performance. To address this gap, our study aims to answer three core questions:
\begin{itemize}\setlength{\itemsep}{0pt}
\item[(1)] How prevalent is attribute imbalance in commonly used vision datasets?
\item[(2)] What is the impact of attribute imbalance on model performance?
\item[(3)] How can attribute imbalance be effectively mitigated?
\end{itemize}

To automatically assess the degree of attribute imbalance in image datasets, two key challenges must be addressed: how to define attributes and how to determine the attributes corresponding to each image. First, based on previous studies \cite{paper1,zhang2024concept}, we define $20$ primary attributes (e.g., color) and their corresponding $300+$ secondary attributes (e.g., black, white). Second, leveraging the CLIP \cite{clip}, we construct a visual attribute dictionary that aligns low-level visual attributes of images with specific textual descriptions, enabling automated attribute annotation for each image. Using this dictionary, we assign the most suitable secondary attribute from each primary attribute category to each image, resulting in a total of $20$ secondary attributes per image. We then compute the frequency of all secondary attributes within each class—the lower the frequency, the higher the scarcity. Based on this, we propose the concept of Compositional Attribute Scarcity (CAS) to comprehensively evaluate the overall attribute scarcity of an individual image. Specifically, for each image, we calculate the scarcity of its contained secondary attributes and sum them to obtain its CAS score.

\begin{figure*}[t]
\begin{center}
\centerline{\includegraphics[width=2\columnwidth]{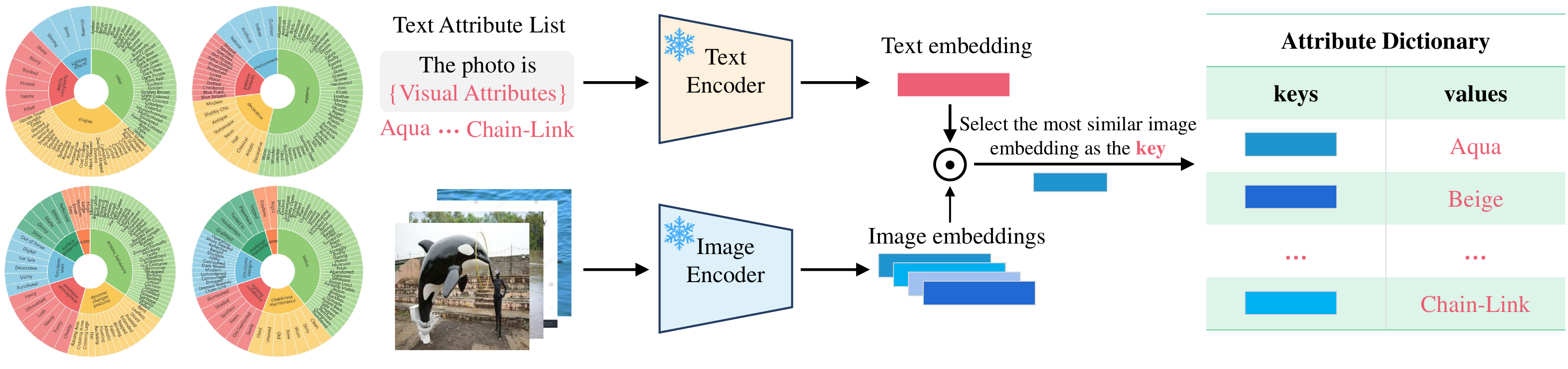}}
\vskip -0.1in
\caption{The left side shows all primary attributes we defined and their corresponding secondary attributes. The right side illustrates the process of constructing the visual attribute dictionary based on CLIP.}
\label{fig1}
\end{center}
\vskip -0.3in
\end{figure*}

Through experiments on $12$ commonly used vision datasets, we reveal that intra-class attribute imbalance and compositional attribute imbalance are pervasive. Furthermore, we systematically analyze how these imbalances affect model performance. Our experimental results demonstrate a consistent pattern: images with higher CAS tend to have lower recognition accuracy. This finding underscores the potential for improving model generalization by addressing attribute imbalance, beyond the improvements achievable by resolving inter-class imbalance alone.

To mitigate attribute imbalance, we propose a novel sampling adjustment strategy for data augmentation. Specifically, we adjust the sampling probability of each image based on its compositional attribute scarcity, with rarer images being sampled more frequently. This adjustment increases the representation of rare attributes in augmented datasets, enabling data augmentation methods to generate more samples emphasizing rare attributes (e.g., white dogs). As a result, the proposed method facilitates better learning of diverse intra-class attributes. Notably, our method introduces no additional computational overhead and requires only a simple modification of the sampling strategy, making it easily integrated into existing frameworks.
The key contributions of this work are as follows:
\vspace{-10pt}
\begin{itemize}\setlength{\itemsep}{0pt}
\item[(1)] A Visual Attribute Framework: We define a comprehensive visual attribute framework encompassing $20$ primary attributes and over $300$ secondary attributes (Section \ref{3.1}). We also propose a CLIP-based visual attribute dictionary to automate the evaluation of attribute imbalance, revealing its widespread prevalence in general vision datasets (Section \ref{3.2}).
\item[(2)] Impact Analysis of Attribute Imbalance: We reveal the significant impact of attribute imbalance on model performance. Specifically, images with higher CAS exhibit lower recognition accuracy, highlighting the necessity and importance of addressing intra-class attribute imbalance (Sections \ref{3.3} and \ref{3.4}).
\item[(3)] We propose a sampling adjustment method based on CAS. This method, requiring only a custom sampler, integrates seamlessly with existing data augmentation frameworks (Section \ref{sec4}). Experiments on $12$ benchmark datasets demonstrate the effectiveness and generalizability of the proposed approach (Section \ref{sec5}).
\end{itemize}

\section{Related Work}

\subsection{Long-tailed image recognition}

In practice, the dataset usually tends to follow a long-tailed distribution, which leads to models with very large variances in performance on each class. It should be noted that most researchers default to the main motivation for long-tail visual recognition is that classes with few samples are always weak classes. Therefore, numerous methods have been proposed to improve the performance of the model on tail classes. \cite{paper58} divides these methods into three fields, namely class rebalancing \cite{paper27,paper4,paper3,paper59,paper60,paper61,paper62,paper63, paper31,paper64,paper65,paper1}, information augmentation \cite{paperma,paperma1,paper10,paper41,paper8,paper72,paper73,paper74,paper71}, and module improvement \cite{paper13,paper78,paper5,paper31,paper77,paper7,paper34}. Unlike the above, \cite{paper27} and \cite{paper28} observe that the number of samples in the class does not exactly show a positive correlation with the accuracy, and the accuracy of some tail classes is even higher than the accuracy of the head class. Therefore, they propose to use other measures to gauge the learning difficulty of the classes rather than relying on the sample number alone.

\vspace{-4pt}
\subsection{Discussion on Intra-Class Imbalance}

The fundamental goal of exploring intra-class imbalance is to identify factors that cause differences in recognition performance among samples within the same class, thereby enabling targeted model improvements. \cite{liu2021handling} attempted to define an imbalanced distribution of learning difficulty within a class, where learning difficulty is determined by the model's prediction confidence. However, prediction confidence varies across different models, leading to inconsistent quantification of learning difficulty, which lacks reproducibility, transparency, and interpretability. \cite{tang2022invariant} proposed investigating the long-tail distribution of attributes within a class but only provided qualitative analyses of how attribute imbalance might negatively affect model performance.

In practical applications, \cite{doonan2025handling} addressed the imbalanced distribution of plant traits in wheat recognition by applying weighted point cloud sampling to increase the proportion of rare plant traits. Similarly, \cite{yang2024defect} focused on generating data for specific defect types in industrial defect detection to balance subclass distributions, effectively improving defect detection accuracy. \cite{zhou2023global} explored the relationship between noise interference and camera angle imbalance with segmentation performance in medical image segmentation tasks. However, these studies are limited to specific domains and lack generalizability. 

To date, no research has systematically defined general visual attributes, thoroughly investigated the prevalence of attribute imbalance, or examined whether its negative impact on models warrants widespread attention from researchers.

\section{Attribute Imbalance}
\label{Attribute_Imbalance}

In this section, we first systematically define the visual attributes of images. Then, we propose using CLIP to construct a visual attribute dictionary, enabling automatic evaluation of image attributes. Finally, we reveal the prevalence of attribute imbalance and compositional attribute imbalance across $12$ commonly used visual datasets and analyze their impact on model performance.

\subsection{Definition of Visual Attributes}
\label{3.1}

Visual attributes refer to the fundamental characteristics that constitute an image, such as color, texture, and shape. These attributes not only define the visual appearance of an image but also play a critical role in the representation learning process of deep learning models.
In this study, we define visual attributes based on a comprehensive analysis of prior research \cite{paper1_1,paper2_2} and practical insights. These attributes are categorized into $20$ primary attributes (e.g., color, material, shape, size) and over $300$ secondary attributes (e.g., ``black" and ``white" under color). Figure \ref{fig1} illustrates all primary and secondary attributes. This hierarchical design ensures both the comprehensiveness and granularity of attribute definitions.

\begin{figure*}[t]
\begin{center}
\centerline{\includegraphics[width=2\columnwidth]{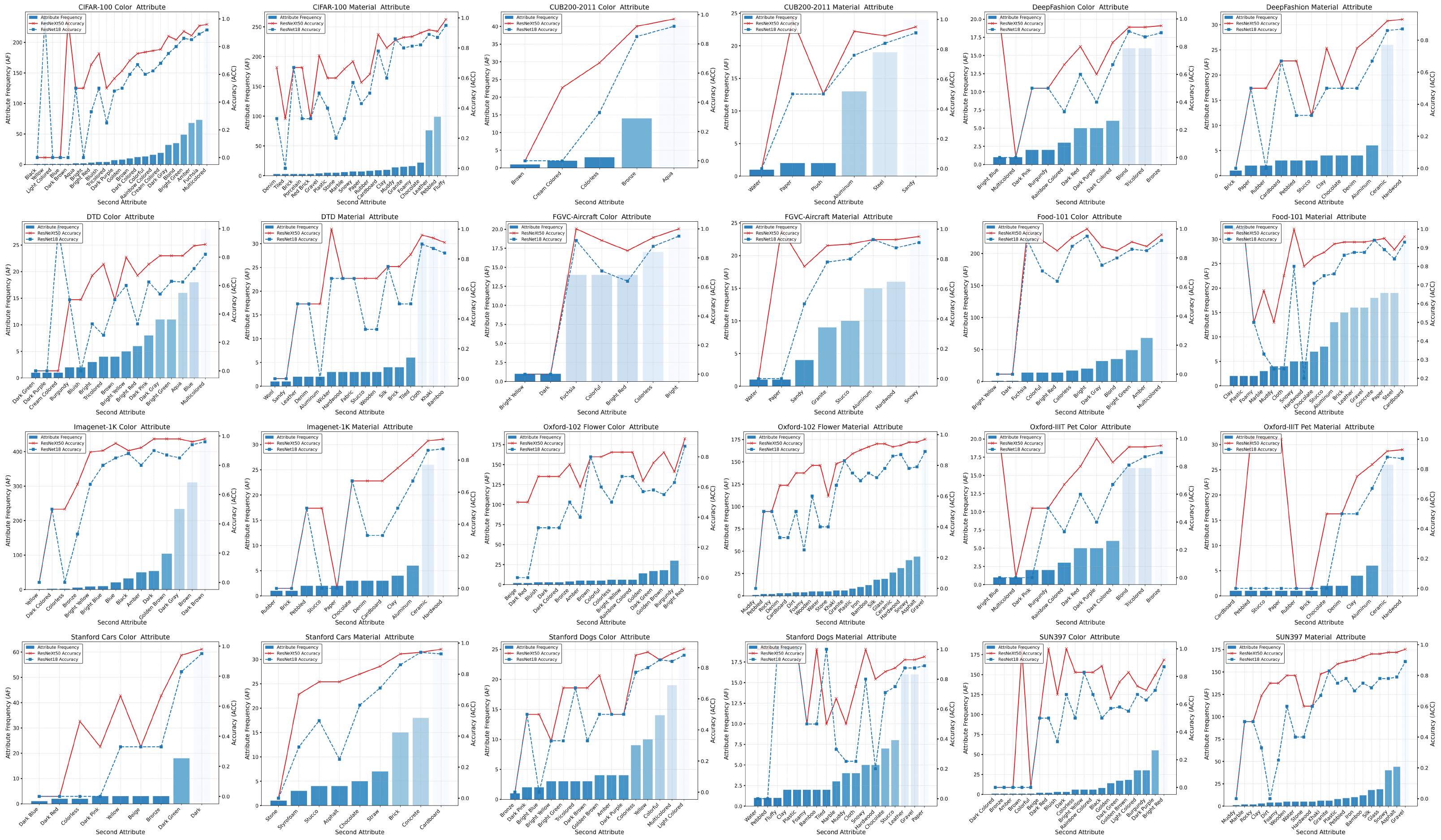}}
\vskip -0.1in
\caption{The distribution of secondary attributes under color and material categories across $12$ visual benchmark datasets, along with the performance of ResNet-18 and ResNeXt-50 on each secondary attribute.}
\label{fig2}
\end{center}
\vskip -0.3in
\end{figure*}

\subsection{Constructing a Visual Attribute Dictionary}
\label{3.2}

To enable the automated evaluation of attribute distributions in image datasets, we leverage the CLIP model to construct a visual attribute dictionary on ImageNet-21k. As shown in Figure \ref{fig1}, we first organize all secondary attributes into a textual attribute list, such as ``The photo is Brown," and generate corresponding text embeddings. Next, we calculate the similarity between each text embedding and the image embeddings, matching the most similar image embedding to the respective text attribute.
The matched image embeddings serve as the keys in the visual attribute dictionary, while the corresponding text attributes serve as the values. To query the visual attributes of a given image, its embedding is extracted and compared with the dictionary keys using cosine similarity. The value corresponding to the key with the highest similarity score is then returned as the predicted attribute for the image.

\subsection{Single-Attribute Imbalance}
\label{3.3}

At the single-attribute level, the imbalance manifests as certain attributes (e.g., ``black") dominating a large proportion of the dataset, while other attributes (e.g., ``purple") are represented by only a few samples. We conducted a systematic analysis of $12$ commonly used visual datasets, including ImageNet and CIFAR-100. Using the visual attribute dictionary, we calculated the distribution of different attributes in each dataset and quantified the degree of attribute imbalance. As shown in Figure \ref{fig2}, the distribution of secondary attributes under each primary attribute typically exhibits a long-tailed pattern, with a large number of low-frequency attributes having significantly fewer samples than high-frequency attributes.

To investigate the impact of attribute frequency on model performance, we first trained standard ResNet-18 and ResNet-50 models on each dataset. Within each category, for each primary attribute, we divided the samples into subsets based on their associated secondary attributes and evaluated the recognition accuracy of both models on each subset. The experimental results, shown in Figure \ref{fig2}, reveal that samples with higher-frequency attributes generally achieve higher and more stable recognition accuracy. Conversely, samples with low-frequency attributes do not consistently exhibit the expected low recognition accuracy. Merely analyzing single-attribute imbalance is insufficient to explain this phenomenon. We hypothesize that the rarity of one type of secondary attribute in an image does not necessarily imply the rarity of other secondary attributes (e.g., a rare color may coexist with a common shape).

\subsection{Compositional Attribute Imbalance}
\label{3.4}

In the analysis of single-attribute imbalance, we observed that samples with high-frequency attributes are more likely to be correctly identified by the model. However, merely relying on single-attribute statistics cannot fully explain the model's performance on low-frequency attribute samples. Considering that an image often contains multiple visual attributes, we further introduce the concept of compositional attributes to explore the impact of multi-attribute scarcity on model performance.

Compositional attributes refer to the specific combination of multiple ($20$ in this study) primary attributes in an image, such as \{Blue, Metallic, Round, …\} or \{Red, Wooden, Square, …\}. These combinations not only describe the visual characteristics of an image but also capture the interrelationships between attributes. However, the free combinations of attributes are not uniformly distributed in datasets, and many compositional attributes are extremely scarce in the training data. Such scarcity may lead to significantly degraded model performance on images with these rare compositional attributes.
We define \textbf{Compositional Attribute Scarcity (CAS)} as follows:
\begin{itemize}\setlength{\itemsep}{0pt}
\item[(1)] For each primary attribute, calculate the frequency of its secondary attributes and rank them in descending order of frequency.
\item[(2)] The scarcity of each secondary attribute is indicated by the ranked position, the lower the rank, the rarer it is.
\item[(3)] The CAS of an image is calculated as the sum of the scarcity ranks of its $20$ secondary attributes.
\end{itemize}

\begin{figure}[t]
\vskip -0.15in
\begin{center}
\centerline{\includegraphics[width=1\columnwidth]{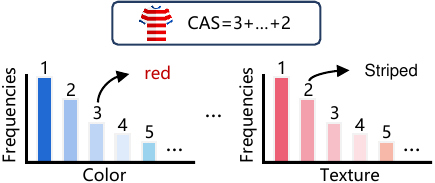}}
\vskip -0.1in
\caption{Illustration of the computation process for image ompositional attribute scarcity (CAS).}
\label{fig3}
\end{center}
\vskip -0.25in
\end{figure}

\begin{figure*}[t]
\begin{center}
\centerline{\includegraphics[width=2\columnwidth]{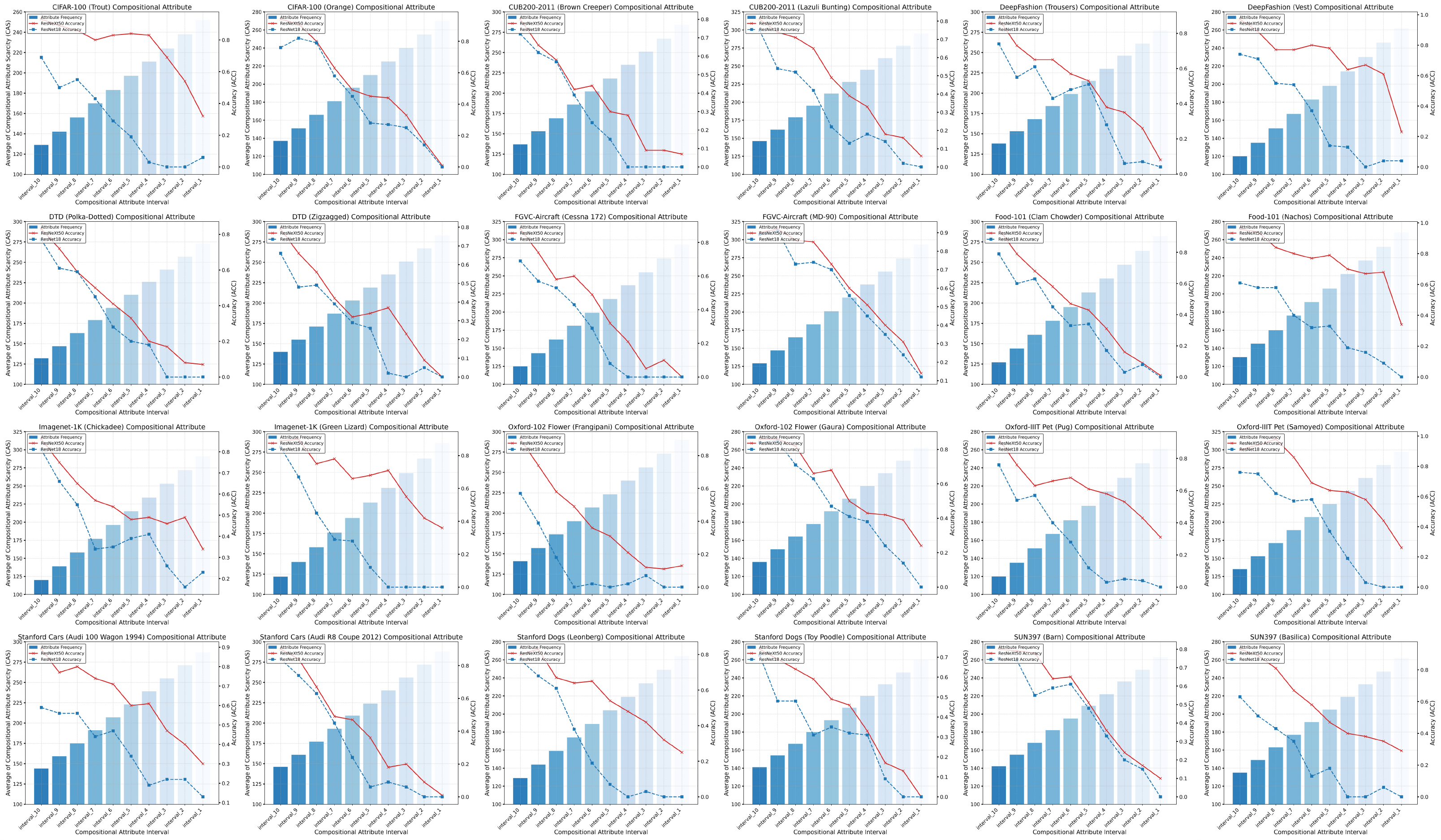}}
\vskip -0.05in
\caption{The long-tailed distribution of sample composite attribute sparsity across certain categories in $12$ visual benchmark datasets, along with the performance of ResNet-18 and ResNeXt-50 across different compositional attribute scarcity (CAS) intervals. The horizontal axis represents $10$ evenly divided intervals based on different CAS values, increasing from left to right. The left vertical axis indicates the average compositional attribute scarcity of all samples within each interval.}
\label{fig4}
\end{center}
\vskip -0.2in
\end{figure*}

Figure \ref{fig3} illustrates the process of calculating the CAS of an image.
We further investigate the impact of compositional attribute scarcity on model performance across $12$ datasets using ResNet-18 and ResNet-50. Samples were divided into subsets based on their CAS values, and classification accuracy was evaluated for each subset. The experimental results, shown in Figure \ref{fig4}, reveal the following:
\begin{itemize}\setlength{\itemsep}{0pt}
\item[(1)] Compositional attribute imbalance is pervasive, with many attribute combinations represented by only a few samples in the entire dataset.
\item[(2)] As Compositional Attribute Scarcity increases, model performance gradually deteriorates.
\end{itemize}
It is evident that samples with high compositional attribute scarcity are often under-learned. To address this issue, we propose a simple yet effective solution to mitigate the impact of compositional attribute imbalance.

\section{Leveraging Compositional Attribute Scarcity to Guide Data Augmentation}
\label{sec4}

To mitigate the negative impact of visual attribute imbalance on model performance, we propose a simple yet effective solution. The core idea is to adjust the sampling probability during data augmentation based on a sample's compositional attribute scarcity, thereby generating more samples with rare attributes. This approach enhances the model's representation capability for these underrepresented attributes. To amplify the differences in scarcity among samples, we introduce a power transformation to nonlinearly enhance the scarcity values. In practice, our method can be seamlessly integrated with existing data augmentation techniques by customizing the sampler.

\subsection{Sampling Strategy Based on CAS}
\label{4.1}

Assume the total number of samples is \( M \), and the compositional attribute scarcity of sample \( i \) is \( r_i \). To enhance the differentiation of scarcity, we apply a power transformation to \( r_i \): $r_i' = r_i^b,$
where \( b \) is the power parameter controlling the degree of nonlinear amplification. When \( b > 1 \), the differentiation of high-scarcity samples is significantly increased. Our empirical studies recommend setting \( b \) to $1.2$ (see Section \ref{sec5.4}). Based on the transformed scarcity \( r_i' \), the sampling probability for each sample is defined as:  
$$p_i = \frac{r_i'}{\sum_{k=1}^M r_k'}.$$

\setlength{\textfloatsep}{23pt}
\begin{algorithm}[t]
\SetKwInOut{Input}{Input}
\SetKwInOut{Output}{Output}

\caption{Enhancing CutMix with CAS}
\label{alg1}
\Input{
    Dataset $\mathcal{D} = \{(x_i, y_i)\}_{i=1}^N$, \\
    Combination rarity scores $\{r_i\}_{i=1}^N$, \\
    CutMix parameter $\alpha > 0$, \\
    Training epochs $T$, Scaling factor $\beta > 0$.
}
\Output{Trained model $\mathcal{M}$.}

\BlankLine
\textbf{Step 1: Compute Sampling Weights}\\
\ForEach{$r_i$ in $\{r_1, r_2, \dots, r_N\}$}{
    Compute weight $w_i \gets r_i^\beta$\;
}
Define weights vector $\mathbf{w} \gets \{w_1, w_2, \dots, w_N\}$\;

\BlankLine
\textbf{Step 2: Initialize Weighted Sampler}\\
Initialize sampler $\mathcal{S}$ with weights $\mathbf{w}$ using \texttt{WeightedRandomSampler}\;

\BlankLine
\textbf{Step 3: Training with CutMix}\\
\For{$t = 1$ \textbf{to} $T$}{
    Sample a batch $\mathcal{B}$ from $\mathcal{D}$ using sampler $\mathcal{S}$\;
    \ForEach{pair $(x_i, y_i)$ and $(x_j, y_j)$ in $\mathcal{B}$}{
        Sample $\lambda \sim \text{Beta}(\alpha, \alpha)$\;
        Compute CutMix bounding box $(B)$\;
        Generate CutMix samples: \\
        $x_{mix} \gets x_i \cdot M + x_j \cdot (1 - M)$, where $M$ is the binary mask defined by $B$\;
        $y_{mix} \gets \lambda y_i + (1 - \lambda) y_j$\;
    }
    Perform forward and backward propagation on CutMix samples\;
    Update model $\mathcal{M}$\;
}

\BlankLine
\Return{Trained model $\mathcal{M}$.}
\end{algorithm}

\renewcommand{\arraystretch}{1.1}
\begin{table}[b]
\vspace{-5pt}
\centering
\resizebox{\linewidth}{!}{ 
\begin{tabular}{c|l|l|l}
\hline \hline
\multicolumn{1}{l}{} & \multicolumn{1}{|c|}{ImageNet-1K} & \multicolumn{1}{c|}{+CutMix} & \multicolumn{1}{c}{+Our Method} \\ \hline
Mean                   & 124.6                            & 120.4                        & 129.8                            \\ \hline
Standard Deviation     & 37.6                             & 36.8                          & 29.3                             \\ \hline \hline
\end{tabular}
}
\caption{Comparison of sample CAS statistics on ImageNet-1K. Our method significantly reduces the standard deviation of CAS, indicating a more balanced and less dispersed distribution of compositional attributes.}
\label{tab1}
\vspace{-5pt}
\end{table}

\vspace{-10pt}
\subsection{Seamless Integration with Data Augmentation}
\label{4.2}

During training, we first compute the compositional attribute scarcity and sampling probability for each sample. These probabilities are then used to customize the sampler. Subsequently, data augmentation techniques (e.g., CutMix, FMix, SaliencyMix) are applied to preferentially generate more samples with rare attributes. Algorithm \ref{alg1} provides the implementation details using CutMix as an example. 

Furthermore, Table \ref{tab1} compares the level of compositional attribute imbalance in the dataset before and after applying our method. The results show that using only standard data augmentation strategies yields little improvement in reducing attribute imbalance within the dataset.

\section{Empirical Study}
\label{sec5}

\subsection{Datasets}
To comprehensively evaluate the performance of the proposed method, twelve diverse image classification datasets were selected. These datasets encompass tasks ranging from large-scale image classification to fine-grained classification, which effectively validate the model's performance across various scenarios. The \texttt{ImageNet-1K} \cite{deng2009imagenet} dataset contains $1.2$ million training images and $50,000$ validation images across $1,000$ categories. The \texttt{CIFAR-100} \cite{krizhevsky2009learning} dataset includes $50,000$ training images and $10,000$ test images spanning $100$ categories. The \texttt{Oxford-IIIT Pet} \cite{parkhi2012cats} dataset comprises $37$ pet categories with $7,349$ images. The \texttt{Stanford Dogs} \cite{khosla2011novel} dataset contains $20$ dog breeds with a total of $20,580$ images. The \texttt{DTD (Describable Textures Dataset)} \cite{cimpoi2014describing} includes $47$ texture categories with $1,880$ images. The \texttt{Oxford-102 Flower} \cite{nilsback2008automated} dataset contains $102$ flower categories with $8,189$ images. The \texttt{Food-101} \cite{bossard2014food} dataset covers $101$ food categories with a total of $101,000$ images. The \texttt{Stanford Cars} \cite{krause20133d} dataset includes $196$ car categories with $16,185$ images. The \texttt{FGVC-Aircraft} \cite{maji2013fine} dataset contains $100$ aircraft categories with $10,000$ images. The \texttt{SUN397} \cite{xiao2010sun} dataset features $397$ scene categories with $108,754$ images. The \texttt{DeepFashion} \cite{liu2016deepfashion} dataset consists of $50$ clothing categories with $50,000$ images. Finally, the \texttt{CUB200-2011} \cite{wah2011caltech} dataset includes $200$ bird species categories with $11,788$ images.
Through comprehensive testing across these datasets, we can thoroughly assess the model's performance in a variety of tasks and environments.

\renewcommand{\arraystretch}{0.97}
\begin{table*}[!t]
    \centering
    \begin{subtable}[t]{0.49\linewidth}
        \centering
        \resizebox{\linewidth}{!}
        {
        \begin{tabular}{c|cccc|cccc}
            \hline\noalign{\smallskip}
            \multicolumn{9}{c}{CIFAR-100}\\
            \hline
            \multirow{2}{*}{Method} & \multicolumn{4}{c|}{ResNet18} & \multicolumn{4}{c}{ResNeXt50} \\  
            \cline{2-9} & low & middle & high & all & low & middle & high & all \\  
            \hline
            CutMix    & 68.31 & 49.65 & 43.05 & 54.57 & 71.54 & 51.73 & 44.62 & 58.07\\
            \cellcolor{lightgray!30}CutMix+weight & \cellcolor{lightgray!30}68.96 & \cellcolor{lightgray!30}50.32 & \cellcolor{lightgray!30}43.94 & \cellcolor{lightgray!30}55.28 & \cellcolor{lightgray!30}71.85 & \cellcolor{lightgray!30}52.36& \cellcolor{lightgray!30}45.6 & \cellcolor{lightgray!30}58.52 \\
            $\Delta$ & \incre{+0.65} & \incre{+0.67} & \incre{+0.89}& \incre{+0.71} & \incre{+0.31}& \incre{+0.63} & \incre{+0.98}& \incre{+0.45} \\
            \hline
            FMix    & 64.15 & 52.36 & 33.58 & 51.30 & 69.44& 43.25 & 38.45 & 52.98 \\
            \cellcolor{lightgray!30}FMix+weight & \cellcolor{lightgray!30}66.73 & \cellcolor{lightgray!30}56.63 & \cellcolor{lightgray!30}39.38 & \cellcolor{lightgray!30}52.85 & \cellcolor{lightgray!30}73.69 & \cellcolor{lightgray!30}45.59& \cellcolor{lightgray!30}42.30 & \cellcolor{lightgray!30}55.74 \\
            $\Delta$ & \incre{+2.58} & \incre{+4.27} & \incre{+5.80}& \incre{+1.55} & \incre{+4.25}& \incre{+2.34} & \incre{+3.85}& \incre{+2.76} \\
            \hline
            SaliencyMix    & 79.24 & 64.28 & 41.73 & 57.25 & 72.65& 70.28 & 43.26& 59.65 \\
            \cellcolor{lightgray!30}SaliencyMix+weight & \cellcolor{lightgray!30}79.56 & \cellcolor{lightgray!30}65.86 & \cellcolor{lightgray!30}44.15 & \cellcolor{lightgray!30}57.88 & \cellcolor{lightgray!30}73.41 & \cellcolor{lightgray!30}71.21 & \cellcolor{lightgray!30}45.02 & \cellcolor{lightgray!30}59.94 \\
            $\Delta$ & \incre{+0.32} & \incre{+1.58} & \incre{+2.42}& \incre{+0.63} & \incre{+0.76}& \incre{+0.93} & \incre{+1.76}& \incre{+0.29} \\
            \hline
        \end{tabular}
        }
        \caption{CIFAR-100}
    \end{subtable}
    \begin{subtable}[t]{0.49\linewidth}
        \centering
        \resizebox{\linewidth}{!}
        {
        \begin{tabular}{c|cccc|cccc}
            \hline\noalign{\smallskip}
            \multicolumn{9}{c}{Imagenet-1k}\\
            \hline
            \multirow{2}{*}{Method} & \multicolumn{4}{c|}{ResNet18} & \multicolumn{4}{c}{ResNeXt50} \\  
            \cline{2-9} & low & middle & high & all & low & middle & high & all \\  
            \hline
            CutMix    & 68.24 & 47.26 & 21.68 & 48.36 & 76.30 & 71.14 & 30.25 & 49.78 \\
            \cellcolor{lightgray!30}CutMix+weight & \cellcolor{lightgray!30}68.41 & \cellcolor{lightgray!30}48.22 & \cellcolor{lightgray!30}23.26 & \cellcolor{lightgray!30}49.04 & \cellcolor{lightgray!30}77.28 & \cellcolor{lightgray!30}72.21& \cellcolor{lightgray!30}33.79 & \cellcolor{lightgray!30}50.96 \\
            $\Delta$ & \incre{+0.17} & \incre{+0.96} & \incre{+1.58}& \incre{+0.68} & \incre{+0.98}& \incre{+1.07} & \incre{+3.54}& \incre{+1.18} \\
            \hline
            FMix    & 58.72 & 49.36 & 38.25 & 47.26 & 62.95 & 56.31 & 41.36& 49.78 \\
            \cellcolor{lightgray!30}FMix+weight & \cellcolor{lightgray!30}59.07 & \cellcolor{lightgray!30}50.83 & \cellcolor{lightgray!30}40.00 & \cellcolor{lightgray!30}48.31 & \cellcolor{lightgray!30}63.59 & \cellcolor{lightgray!30}58.18& \cellcolor{lightgray!30}43.97 & \cellcolor{lightgray!30}51.36 \\
            $\Delta$ & \incre{+0.35} & \incre{+1.47} & \incre{+1.75}& \incre{+1.05} & \incre{+0.64}& \incre{+1.87} & \incre{+2.61}& \incre{+1.58} \\
            \hline
            SaliencyMix    & 61.54 & 46.57 & 40.38 & 47.10 & 58.32 & 49.62 & 43.28 & 47.35 \\
            \cellcolor{lightgray!30}SaliencyMix+weight & \cellcolor{lightgray!30}63.22 & \cellcolor{lightgray!30}48.55 & \cellcolor{lightgray!30}42.36 & \cellcolor{lightgray!30}49.78 & \cellcolor{lightgray!30}61.00 & \cellcolor{lightgray!30}52.68& \cellcolor{lightgray!30}48.17 & \cellcolor{lightgray!30}50.42 \\
            $\Delta$ & \incre{+1.68} & \incre{+1.98} & \incre{+2.92}& \incre{+2.68} & \incre{+2.68}& \incre{+3.06} & \incre{+4.89}& \incre{+3.07} \\
            \hline
        \end{tabular}
        }
        \caption{Imagenet-1k}
    \end{subtable}
    \begin{subtable}[t]{0.49\linewidth}
        \centering
        \resizebox{\linewidth}{!}
        {
        \begin{tabular}{c|cccc|cccc}
            \hline\noalign{\smallskip}
            \multicolumn{9}{c}{DTD}\\
            \hline
            \multirow{2}{*}{Method} & \multicolumn{4}{c|}{ResNet18} & \multicolumn{4}{c}{ResNeXt50} \\  
            \cline{2-9} & low & middle & high & all & low & middle & high & all \\  
            \hline
            CutMix    & 95.36 & 91.74 & 85.55 & 91.63 & 95.22 & 94.36 & 86.32 & 99.61 \\
            \cellcolor{lightgray!30}CutMix+weight & \cellcolor{lightgray!30}95.79 & \cellcolor{lightgray!30}92.33 & \cellcolor{lightgray!30}88.43 & \cellcolor{lightgray!30}93.01 & \cellcolor{lightgray!30}95.98 & \cellcolor{lightgray!30}95.05 & \cellcolor{lightgray!30}87.80 & \cellcolor{lightgray!30}99.98 \\
            $\Delta$ & \incre{+0.43} & \incre{+0.59} & \incre{+2.88}& \incre{+1.38} & \incre{+0.17}& \incre{+0.69} & \incre{+1.51}& \incre{+0.37} \\
            \hline
            FMix    & 93.66 & 86.20 & 79.65 & 88.61 & 94.62 & 86.31 & 70.89 & 99.56 \\
            \cellcolor{lightgray!30}FMix+weight & \cellcolor{lightgray!30}95.58 & \cellcolor{lightgray!30}87.89 & \cellcolor{lightgray!30}81.02 & \cellcolor{lightgray!30}88.76 & \cellcolor{lightgray!30}95.38 & \cellcolor{lightgray!30}87.99 & \cellcolor{lightgray!30}72.83 & \cellcolor{lightgray!30}99.65 \\
            $\Delta$ & \incre{+1.92} & \incre{+1.69} & \incre{+1.37}& \incre{+0.15} & \incre{+0.76}& \incre{+1.68} & \incre{+1.94}& \incre{0.09} \\
            \hline
            SaliencyMix    & 92.99 & 81.45 & 69.02 & 89.33 & 91.36 & 79.26 & 65.35& 98.14 \\
            \cellcolor{lightgray!30}SaliencyMix+weight & \cellcolor{lightgray!30}93.45 & \cellcolor{lightgray!30}82.99 & \cellcolor{lightgray!30}70.67 & \cellcolor{lightgray!30}90.65 & \cellcolor{lightgray!30}91.68 & \cellcolor{lightgray!30}80.04& \cellcolor{lightgray!30}69.03 & \cellcolor{lightgray!30}99.99 \\
            $\Delta$ & \incre{+0.46} & \incre{+1.54} & \incre{+1.65}& \incre{+1.32} & \incre{+0.32}& \incre{+0.78} & \incre{+3.68}& \incre{+1.85} \\
            \hline
        \end{tabular}
        }
        \caption{DTD}
    \end{subtable}
    \begin{subtable}[t]{0.49\linewidth}
        \centering
        \resizebox{\linewidth}{!}
        {
        \begin{tabular}{c|cccc|cccc}
            \hline\noalign{\smallskip}
            \multicolumn{9}{c}{FGVC-Aircraft}\\
            \hline
            \multirow{2}{*}{Method} & \multicolumn{4}{c|}{ResNet18} & \multicolumn{4}{c}{ResNeXt50} \\  
            \cline{2-9} & low & middle & high & all & low & middle & high & all \\  
            \hline
            CutMix    & 80.77 & 71.36 & 60.98& 76.25 & 89.31& 79.36 & 61.28& 86.47 \\
            \cellcolor{lightgray!30}CutMix+weight & \cellcolor{lightgray!30}81.46 & \cellcolor{lightgray!30}73.24 & \cellcolor{lightgray!30}63.60 & \cellcolor{lightgray!30}77.29 & \cellcolor{lightgray!30}89.42 & \cellcolor{lightgray!30}80.59 & \cellcolor{lightgray!30}63.13 & \cellcolor{lightgray!30}87.43 \\
            $\Delta$ & \incre{+0.69} & \incre{+1.88} & \incre{+2.62}& \incre{+1.04} & \incre{+0.11}& \incre{+1.23} & \incre{+1.85}& \incre{+0.96} \\
            \hline
            FMix    & 81.33 & 71.02 & 69.35& 74.25 & 89.36 & 78.86 & 70.25& 82.72 \\
            \cellcolor{lightgray!30}FMix+weight & \cellcolor{lightgray!30}82.55 & \cellcolor{lightgray!30}73.38 & \cellcolor{lightgray!30}72.05 & \cellcolor{lightgray!30}75.89 & \cellcolor{lightgray!30}89.61 & \cellcolor{lightgray!30}80.54& \cellcolor{lightgray!30}73.19 & \cellcolor{lightgray!30}84.58 \\
            $\Delta$ & \incre{+1.22} & \incre{+2.36} & \incre{+2.70}& \incre{+1.64} & \incre{+0.25}& \incre{+1.68} & \incre{+2.94}& \incre{+1.86} \\
            \hline
            SaliencyMix    & 85.56 & 78.89 & 68.36& 72.32 & 91.36& 81.22 & 79.69& 85.21 \\
            \cellcolor{lightgray!30}SaliencyMix+weight & \cellcolor{lightgray!30}85.87 & \cellcolor{lightgray!30}79.87 & \cellcolor{lightgray!30}70.33 & \cellcolor{lightgray!30}72.87 & \cellcolor{lightgray!30}92.02 & \cellcolor{lightgray!30}82.56& \cellcolor{lightgray!30}81.21 & \cellcolor{lightgray!30}85.48 \\
            $\Delta$ & \incre{+0.31} & \incre{+0.98} & \incre{+1.97}& \incre{+0.64} & \incre{+0.66}& \incre{+1.34} & \incre{+1.52}& \incre{+0.27} \\
            \hline
        \end{tabular}
        }
        \caption{FGVC-Aircraft}
    \end{subtable}
    \begin{subtable}[t]{0.49\linewidth}
        \centering
        \resizebox{\linewidth}{!}
        {
        \begin{tabular}{c|cccc|cccc}
            \hline\noalign{\smallskip}
            \multicolumn{9}{c}{CUB200-2011}\\
            \hline
            \multirow{2}{*}{Method} & \multicolumn{4}{c|}{ResNet18} & \multicolumn{4}{c}{ResNeXt50} \\  
            \cline{2-9} & low & middle & high & all & low & middle & high & all \\  
            \hline
            CutMix    & 90.86 & 91.68 & 86.35 & 92.36 & 98.35 & 89.63 & 75.36& 94.31 \\
            \cellcolor{lightgray!30}CutMix+weight & \cellcolor{lightgray!30}91.16 & \cellcolor{lightgray!30}92.66 & \cellcolor{lightgray!30}87.93 & \cellcolor{lightgray!30}93.32 & \cellcolor{lightgray!30}99.69 & \cellcolor{lightgray!30}92.31& \cellcolor{lightgray!30}78.34 & \cellcolor{lightgray!30}95.83 \\
            $\Delta$ & \incre{+0.33} & \incre{+0.98} & \incre{+1.58}& \incre{+0.96} & \incre{+1.34}& \incre{+2.68} & \incre{+2.98}& \incre{+1.52} \\
            \hline
            FMix    & 92.88 & 89.36 & 90.58& 90.32 & 91.31 & 89.56 & 86.77 & 96.79 \\
            \cellcolor{lightgray!30}FMix+weight & \cellcolor{lightgray!30}93.43 & \cellcolor{lightgray!30}89.68 & \cellcolor{lightgray!30}92.26 & \cellcolor{lightgray!30}91.00 & \cellcolor{lightgray!30}91.64 & \cellcolor{lightgray!30}90.92& \cellcolor{lightgray!30}86.96 & \cellcolor{lightgray!30}96.97 \\
            $\Delta$ & \incre{+0.55} & \incre{+0.32} & \incre{+1.68}& \incre{+0.68} & \incre{+0.33}& \incre{+1.36} & \incre{+1.25}& \incre{+0.18} \\
            \hline
            SaliencyMix    & 90.05 & 91.02 & 89.25 & 93.68 & 92.86 & 95.44 & 89.65& 96.59 \\
            \cellcolor{lightgray!30}SaliencyMix+weight & \cellcolor{lightgray!30}90.70 & \cellcolor{lightgray!30}91.68 & \cellcolor{lightgray!30}90.50 & \cellcolor{lightgray!30}94.00 & \cellcolor{lightgray!30}93.08 & \cellcolor{lightgray!30}95.89& \cellcolor{lightgray!30}90.67 & \cellcolor{lightgray!30}97.05 \\
            $\Delta$ & \incre{+0.65} & \incre{+0.66} & \incre{+1.25}& \incre{+0.32} & \incre{+0.22}& \incre{+0.45} & \incre{+1.02}& \incre{+0.46} \\
            \hline
        \end{tabular}
        }
        \caption{CUB200-2011}
    \end{subtable}
    \begin{subtable}[t]{0.49\linewidth}
        \centering
        \resizebox{\linewidth}{!}
        {
        \begin{tabular}{c|cccc|cccc}
            \hline\noalign{\smallskip}
            \multicolumn{9}{c}{Oxford IIIT Pet}\\
            \hline
            \multirow{2}{*}{Method} & \multicolumn{4}{c|}{ResNet18} & \multicolumn{4}{c}{ResNeXt50} \\  
            \cline{2-9} & low & middle & high & all & low & middle & high & all \\  
            \hline
            CutMix    & 95.02 & 89.95 & 83.36& 85.58 & 93.25& 89.95 & 82.63& 87.76 \\
            \cellcolor{lightgray!30}CutMix+weight & \cellcolor{lightgray!30}95.70 & \cellcolor{lightgray!30}91.88 & \cellcolor{lightgray!30}84.98 & \cellcolor{lightgray!30}86.44 & \cellcolor{lightgray!30}93.37 & \cellcolor{lightgray!30}90.27& \cellcolor{lightgray!30}84.31 & \cellcolor{lightgray!30}88.28 \\
            $\Delta$ & \incre{+0.68} & \incre{+1.93} & \incre{+1.62}& \incre{+0.86} & \incre{+0.12}& \incre{+0.32} & \incre{+1.68}& \incre{+0.48} \\
            \hline
            FMix    & 92.25 & 91.58 & 81.16& 84.45 & 93.03& 91.12 & 84.50& 85.68 \\
            \cellcolor{lightgray!30}FMix+weight & \cellcolor{lightgray!30}93.93 & \cellcolor{lightgray!30}92.04 & \cellcolor{lightgray!30}83.15 & \cellcolor{lightgray!30}85.43 & \cellcolor{lightgray!30}93.54 & \cellcolor{lightgray!30}92.74& \cellcolor{lightgray!30}86.38 & \cellcolor{lightgray!30}86.94 \\
            $\Delta$ & \incre{+1.68} & \incre{+0.54} & \incre{+1.99}& \incre{+0.98} & \incre{+0.51}& \incre{+1.62} & \incre{+1.88}& \incre{+1.26} \\
            \hline
            SaliencyMix    & 90.04 & 91.12 & 86.65& 88.32 & 94.60& 89.99 & 87.75& 88.53 \\
            \cellcolor{lightgray!30}SaliencyMix+weight & \cellcolor{lightgray!30}91.66 & \cellcolor{lightgray!30}93.01 & \cellcolor{lightgray!30}88.68 & \cellcolor{lightgray!30}89.52 & \cellcolor{lightgray!30}95.14 & \cellcolor{lightgray!30}90.65& \cellcolor{lightgray!30}88.99 & \cellcolor{lightgray!30}89.92 \\
            $\Delta$ & \incre{+1.62} & \incre{+1.89} & \incre{+2.03}& \incre{+1.20} & \incre{+0.54}& \incre{+0.66} & \incre{+1.24}& \incre{+1.39} \\
            \hline
        \end{tabular}
        }
        \caption{Oxford IIIT Pet}
    \end{subtable}
    \begin{subtable}[t]{0.49\linewidth}
        \centering
        \resizebox{\linewidth}{!}
        {
        \begin{tabular}{c|cccc|cccc}
            \hline\noalign{\smallskip}
            \multicolumn{9}{c}{Stanford Dogs}\\
            \hline
            \multirow{2}{*}{Method} & \multicolumn{4}{c|}{ResNet18} & \multicolumn{4}{c}{ResNeXt50} \\  
            \cline{2-9} & low & middle & high & all & low & middle & high & all \\  
            \hline
            CutMix    & 74.43 & 69.98 & 62.24& 68.00 & 85.53& 71.15 & 62.25& 69.14 \\
            \cellcolor{lightgray!30}CutMix+weight & \cellcolor{lightgray!30}77.11 & \cellcolor{lightgray!30}74.93 & \cellcolor{lightgray!30}65.89 & \cellcolor{lightgray!30}71.37 & \cellcolor{lightgray!30}89.78 & \cellcolor{lightgray!30}77.48& \cellcolor{lightgray!30}69.50 & \cellcolor{lightgray!30}74.57 \\
            $\Delta$ & \incre{+2.68} & \incre{+4.95} & \incre{+3.65}& \incre{+3.37} & \incre{+4.25}& \incre{+6.33} & \incre{+7.25}& \incre{+5.43} \\
            \hline
            FMix    & 69.94 & 71.15 & 60.35& 61.54 & 81.16& 72.25 & 68.87& 74.30 \\
            \cellcolor{lightgray!30}FMix+weight & \cellcolor{lightgray!30}71.56 & \cellcolor{lightgray!30}74.93 & \cellcolor{lightgray!30}64.37 & \cellcolor{lightgray!30}64.2 & \cellcolor{lightgray!30}84.39 & \cellcolor{lightgray!30}78.07& \cellcolor{lightgray!30}74.13 & \cellcolor{lightgray!30}79.13 \\
            $\Delta$ & \incre{+1.62} & \incre{+3.78} & \incre{+4.02}& \incre{+2.66} & \incre{+3.23}& \incre{+5.82} & \incre{+5.26}& \incre{+4.83} \\
            \hline
            SaliencyMix    & 75.56 & 68.89 & 60.04& 66.24 & 81.15& 74.65 & 70.05& 72.42 \\
            \cellcolor{lightgray!30}SaliencyMix+weight & \cellcolor{lightgray!30}77.90 & \cellcolor{lightgray!30}72.22 & \cellcolor{lightgray!30}66.68 & \cellcolor{lightgray!30}71.5 & \cellcolor{lightgray!30}82.80 & \cellcolor{lightgray!30}76.47& \cellcolor{lightgray!30}72.21 & \cellcolor{lightgray!30}74.27 \\
            $\Delta$ & \incre{+2.34} & \incre{+3.33} & \incre{+6.64}& \incre{+5.26} & \incre{+1.65}& \incre{+1.82} & \incre{+2.16}& \incre{+1.85} \\
            \hline
        \end{tabular}
        }
        \caption{Stanford Dogs}
    \end{subtable}
    \begin{subtable}[t]{0.49\linewidth}
        \centering
        \resizebox{\linewidth}{!}
        {
        \begin{tabular}{c|cccc|cccc}
            \hline\noalign{\smallskip}
            \multicolumn{9}{c}{Food-101}\\
            \hline
            \multirow{2}{*}{Method} & \multicolumn{4}{c|}{ResNet18} & \multicolumn{4}{c}{ResNeXt50} \\  
            \cline{2-9} & low & middle & high & all & low & middle & high & all \\  
            \hline
            CutMix    & 86.66 & 90.05 & 81.16& 84.25 & 92.25& 89.99 & 91.16& 87.76 \\
            \cellcolor{lightgray!30}CutMix+weight & \cellcolor{lightgray!30}87.61 & \cellcolor{lightgray!30}91.93 & \cellcolor{lightgray!30}82.92 & \cellcolor{lightgray!30}85.57 & \cellcolor{lightgray!30}92.93 & \cellcolor{lightgray!30}90.65& \cellcolor{lightgray!30}94.57 & \cellcolor{lightgray!30}89.75 \\
            $\Delta$ & \incre{+0.95} & \incre{+1.88} & \incre{+1.76}& \incre{+1.32} & \incre{+0.68}& \incre{+0.66} & \incre{+3.41}& \incre{+1.99} \\
            \hline
            FMix    & 90.02 & 89.95 & 76.65& 81.66 & 91.17& 84.45 & 80.00& 87.85 \\
            \cellcolor{lightgray!30}FMix+weight & \cellcolor{lightgray!30}91.70 & \cellcolor{lightgray!30}90.20 & \cellcolor{lightgray!30}78.19 & \cellcolor{lightgray!30}82.01 & \cellcolor{lightgray!30}92.13 & \cellcolor{lightgray!30}85.79& \cellcolor{lightgray!30}81.66 & \cellcolor{lightgray!30}88.12 \\
            $\Delta$ & \incre{+1.68} & \incre{+0.25} & \incre{+1.54}& \incre{+0.35} & \incre{+0.96}& \incre{+1.34} & \incre{+1.66}& \incre{+0.27} \\
            \hline
            SaliencyMix    & 92.22 & 91.14 & 89.95& 85.54 & 96.65& 91.99 & 88.82& 87.75 \\
            \cellcolor{lightgray!30}SaliencyMix+weight & \cellcolor{lightgray!30}93.07 & \cellcolor{lightgray!30}92.39 & \cellcolor{lightgray!30}91.12 & \cellcolor{lightgray!30}86.17 & \cellcolor{lightgray!30}97.40 & \cellcolor{lightgray!30}93.61& \cellcolor{lightgray!30}90.58 & \cellcolor{lightgray!30}89.75 \\
            $\Delta$ & \incre{+0.85} & \incre{+1.25} & \incre{+1.47}& \incre{+0.63} & \incre{+0.75}& \incre{+1.62} & \incre{+1.76}& \incre{+1.02} \\
            \hline
        \end{tabular}
        }
        \caption{Food-101}
    \end{subtable}
    \begin{subtable}[t]{0.49\linewidth}
        \centering
        \resizebox{\linewidth}{!}
        {
        \begin{tabular}{c|cccc|cccc}
            \hline\noalign{\smallskip}
            \multicolumn{9}{c}{Stanford Cars}\\
            \hline
            \multirow{2}{*}{Method} & \multicolumn{4}{c|}{ResNet18} & \multicolumn{4}{c}{ResNeXt50} \\  
            \cline{2-9} & low & middle & high & all & low & middle & high & all \\  
            \hline
            CutMix    & 84.89 & 81.36 & 80.02& 82.49 &93.33& 91.47 & 82.24& 92.77 \\
            \cellcolor{lightgray!30}CutMix+weight & \cellcolor{lightgray!30}90.25 & \cellcolor{lightgray!30}89.6 & \cellcolor{lightgray!30}86.68 & \cellcolor{lightgray!30}89.57 & \cellcolor{lightgray!30}94.01 & \cellcolor{lightgray!30}92.72& \cellcolor{lightgray!30}83.99 & \cellcolor{lightgray!30}93.86 \\
            $\Delta$ & \incre{+5.36} & \incre{+8.24} & \incre{+6.66}& \incre{+7.08} & \incre{+0.68}& \incre{+1.25} & \incre{+1.75}& \incre{+1.09} \\
            \hline
            FMix    & 81.14 & 82.25 & 70.02& 75.72 &92.26& 91.14 & 89.92& 90.79 \\
            \cellcolor{lightgray!30}FMix+weight & \cellcolor{lightgray!30}85.99 & \cellcolor{lightgray!30}88.02 & \cellcolor{lightgray!30}76.91 & \cellcolor{lightgray!30}86.51 & \cellcolor{lightgray!30}92.91 & \cellcolor{lightgray!30}92.34& \cellcolor{lightgray!30}91.57 & \cellcolor{lightgray!30}92.12 \\
            $\Delta$ & \incre{+4.85} & \incre{+5.77} & \incre{+6.89}& \incre{+10.79} & \incre{+0.65}& \incre{+1.20} & \incre{+1.65}& \incre{+1.33} \\
            \hline
            SaliencyMix    & 92.13 &89.99 & 84.45& 87.39 & 93.36& 91.12 & 84.45& 92.08 \\
            \cellcolor{lightgray!30}SaliencyMix+weight & \cellcolor{lightgray!30}95.37 & \cellcolor{lightgray!30}97.55 & \cellcolor{lightgray!30}91.33 & \cellcolor{lightgray!30}93.93 & \cellcolor{lightgray!30}94.33 & \cellcolor{lightgray!30}92.81& \cellcolor{lightgray!30}86.33 & \cellcolor{lightgray!30}92.96 \\
            $\Delta$ & \incre{+3.24} & \incre{+7.56} & \incre{+6.88}& \incre{+6.54} & \incre{+0.97}& \incre{+1.69} & \incre{+1.88}& \incre{+0.88} \\
            \hline
        \end{tabular}
        }
        \caption{Stanford Cars}
    \end{subtable}
    \begin{subtable}[t]{0.49\linewidth}
        \centering
        \resizebox{\linewidth}{!}
        {
        \begin{tabular}{c|cccc|cccc}
            \hline\noalign{\smallskip}
            \multicolumn{9}{c}{Deep Fashion}\\
            \hline
            \multirow{2}{*}{Method} & \multicolumn{4}{c|}{ResNet18} & \multicolumn{4}{c}{ResNeXt50} \\  
            \cline{2-9} & low & middle & high & all & low & middle & high & all \\  
            \hline
            CutMix    & 89.94 & 91.14 & 84.46& 87.19 & 93.33& 91.25 & 88.86& 89.55 \\
            \cellcolor{lightgray!30}CutMix+weight & \cellcolor{lightgray!30}92.81 & \cellcolor{lightgray!30}96.09 & \cellcolor{lightgray!30}88.81 & \cellcolor{lightgray!30}90.73 & \cellcolor{lightgray!30}94.79 & \cellcolor{lightgray!30}92.82& \cellcolor{lightgray!30}91.54 & \cellcolor{lightgray!30}92.29 \\
            $\Delta$ & \incre{+2.87} & \incre{+4.95} & \incre{+4.35}& \incre{+3.54} & \incre{+1.46}& \incre{+1.57} & \incre{+2.68}& \incre{+1.74} \\
            \hline
            FMix    & 94.44 & 93.36 & 81.23& 87.37 & 92.28& 91.14 & 87.76& 88.95 \\
            \cellcolor{lightgray!30}FMix+weight & \cellcolor{lightgray!30}94.80 & \cellcolor{lightgray!30}94.8 & \cellcolor{lightgray!30}84.10 & \cellcolor{lightgray!30}82.74 & \cellcolor{lightgray!30}93.17 & \cellcolor{lightgray!30}92.48& \cellcolor{lightgray!30}89.34 & \cellcolor{lightgray!30}89.62 \\
            $\Delta$ & \incre{+0.36} & \incre{+1.44} & \incre{+2.87}& \incre{+1.54} & \incre{+0.89}& \incre{+1.34} & \incre{+1.58}& \incre{+0.67} \\
            \hline
            SaliencyMix    & 93.20 & 90.02 & 86.63& 87.24 & 90.03& 91.15 & 86.65& 90.53 \\
            \cellcolor{lightgray!30}SaliencyMix+weight & \cellcolor{lightgray!30}96.88 & \cellcolor{lightgray!30}92.06 & \cellcolor{lightgray!30}88.85 & \cellcolor{lightgray!30}89.81 & \cellcolor{lightgray!30}90.67 & \cellcolor{lightgray!30}92.46& \cellcolor{lightgray!30}87.85 & \cellcolor{lightgray!30}91.85 \\
            $\Delta$ & \incre{+3.68} & \incre{+2.04} & \incre{+2.22}& \incre{+2.57} & \incre{+0.64}& \incre{+1.31} & \incre{+1.20}& \incre{+1.32} \\
            \hline
        \end{tabular}
        }
        \caption{Deep Fashion}
    \end{subtable}
    \begin{subtable}[t]{0.49\linewidth}
        \centering
        \resizebox{\linewidth}{!}
        {
        \begin{tabular}{c|cccc|cccc}
            \hline\noalign{\smallskip}
            \multicolumn{9}{c}{SUN397}\\
            \hline
            \multirow{2}{*}{Method} & \multicolumn{4}{c|}{ResNet18} & \multicolumn{4}{c}{ResNeXt50} \\  
            \cline{2-9} & low & middle & high & all & low & middle & high & all \\  
            \hline
            CutMix    & 78.85 & 73.36 & 68.95& 70.31 & 79.98& 78.89 & 71.62& 76.35 \\
            \cellcolor{lightgray!30}CutMix+weight & \cellcolor{lightgray!30}81.53 & \cellcolor{lightgray!30}76.92 & \cellcolor{lightgray!30}71.26 & \cellcolor{lightgray!30}73.31 & \cellcolor{lightgray!30}81.20 & \cellcolor{lightgray!30}77.21& \cellcolor{lightgray!30}73.58 & \cellcolor{lightgray!30}77.83 \\
            $\Delta$ & \incre{+2.68} & \incre{+3.56} & \incre{+2.31}& \incre{+3.00} & \incre{+1.22}& \incre{+0.86} & \incre{+1.96}& \incre{+1.48} \\
            \hline
            FMix    & 63.34 & 59.98 & 61.04& 60.03 & 73.36& 69.94 & 65.58& 67.76 \\
            \cellcolor{lightgray!30}FMix+weight & \cellcolor{lightgray!30}65.29 & \cellcolor{lightgray!30}62.34 & \cellcolor{lightgray!30}63.74 & \cellcolor{lightgray!30}62.42 & \cellcolor{lightgray!30}74.02 & \cellcolor{lightgray!30}71.15& \cellcolor{lightgray!30}67.26 & \cellcolor{lightgray!30}68.63 \\
            $\Delta$ & \incre{+1.95} & \incre{+2.36} & \incre{+2.70}& \incre{+2.39} & \incre{+0.66}& \incre{+1.21} & \incre{+1.68}& \incre{+0.87} \\
            \hline
            SaliencyMix    & 63.33 & 64.45 & 58.85& 60.91 &69.99& 68.85 & 60.02& 61.99 \\
            \cellcolor{lightgray!30}SaliencyMix+weight & \cellcolor{lightgray!30}65.37 & \cellcolor{lightgray!30}66.13 & \cellcolor{lightgray!30}62.71 & \cellcolor{lightgray!30}64.35 & \cellcolor{lightgray!30}70.32 & \cellcolor{lightgray!30}70.11& \cellcolor{lightgray!30}61.13 & \cellcolor{lightgray!30}63.33 \\
            $\Delta$ & \incre{+2.04} & \incre{+1.68} & \incre{+3.86}& \incre{+3.44} & \incre{+0.33}& \incre{+1.26} & \incre{+1.11}& \incre{+1.34} \\
            \hline
        \end{tabular}
        }
        \caption{SUN397}
    \end{subtable}
    \begin{subtable}[t]{0.49\linewidth}
        \centering
        \resizebox{\linewidth}{!}
        {
        \begin{tabular}{c|cccc|cccc}
            \hline\noalign{\smallskip}
            \multicolumn{9}{c}{Oxford-102 Flower}\\
            \hline
            \multirow{2}{*}{Method} & \multicolumn{4}{c|}{ResNet18} & \multicolumn{4}{c}{ResNeXt50} \\  
            \cline{2-9} & low & middle & high & all & low & middle & high & all \\  
            \hline
            CutMix    & 93.22 & 94.41 & 91.15& 93.89 & 96.63& 95.53 & 90.24& 97.68 \\
            \cellcolor{lightgray!30}CutMix+weight & \cellcolor{lightgray!30}95.91 & \cellcolor{lightgray!30}99.40 & \cellcolor{lightgray!30}96.51 & \cellcolor{lightgray!30}98.88 & \cellcolor{lightgray!30}97.25 & \cellcolor{lightgray!30}96.77& \cellcolor{lightgray!30}92.92 & \cellcolor{lightgray!30}99.33 \\
            $\Delta$ & \incre{+2.69} & \incre{+4.99} & \incre{+5.36}& \incre{+4.99} & \incre{+0.62}& \incre{+1.24} & \incre{+2.68}& \incre{+1.65} \\
            \hline
            FMix    & 93.33 & 92.25 & 86.62& 90.00 & 94.43& 95.56 & 91.12& 94.62 \\
            \cellcolor{lightgray!30}FMix+weight & \cellcolor{lightgray!30}95.49 & \cellcolor{lightgray!30}97.2 & \cellcolor{lightgray!30}90.30 & \cellcolor{lightgray!30}93.21 & \cellcolor{lightgray!30}97.42 & \cellcolor{lightgray!30}99.04& \cellcolor{lightgray!30}95.78 & \cellcolor{lightgray!30}99.51 \\
            $\Delta$ & \incre{+2.16} & \incre{+4.95} & \incre{+3.68}& \incre{+3.21} & \incre{+2.99}& \incre{+3.48} & \incre{+4.66}& \incre{+3.44} \\
            \hline
            SaliencyMix    &93.34 & 90.06 & 89.95& 91.14 & 90.01& 93.32 & 89.95& 94.62 \\
            \cellcolor{lightgray!30}SaliencyMix+weight & \cellcolor{lightgray!30}96.98 & \cellcolor{lightgray!30}90.22 & \cellcolor{lightgray!30}91.63 & \cellcolor{lightgray!30}93.15 & \cellcolor{lightgray!30}93.99 & \cellcolor{lightgray!30}98.98& \cellcolor{lightgray!30}95.61 & \cellcolor{lightgray!30}99.63 \\
            $\Delta$ & \incre{+3.64} & \incre{+0.16} & \incre{+1.68}& \incre{+2.01} & \incre{+3.98}& \incre{+5.66} & \incre{+5.31}& \incre{+5.01} \\
            \hline
        \end{tabular}
        }
        \caption{Oxford-102 Flower}
    \end{subtable}   
    \vspace{-5pt}
    \caption{Evaluation results on $12$ visual benchmark datasets. The overall performance improvement of CutMix, FMix, and SaliencyMix using our method is reported. Additionally, the performance of our method on three different CAS-based subsets is presented.}
    \label{tab2}
    \vspace{-7pt}
\end{table*}

\subsection{Implementation Details}
In this experiment, we employed \texttt{ResNet-18} and \texttt{ResNeXt-50} as the baseline models and configured the hyperparameter $\alpha$ for different data augmentation methods \cite{sumix}. Specifically, $\alpha$ was set to 0.2 for CutMix, FMix, and SaliencyMix, and the mixed hyperparameters were generated by sampling from a Beta($\alpha$, $\alpha$) distribution at each training iteration. The batch size for all models was set to 64, and the initial learning rate was 0.1, with \texttt{cosine annealing} used as the learning rate scheduling strategy \cite{sumix,diffusemix}.

\subsection{Evaluation Metrics}
To comprehensively evaluate the performance of the proposed method, in addition to testing overall classification accuracy \cite{sumix}, we also introduce a sample partitioning strategy based on compositional attribute scarcity (CAS) to further investigate the model's performance at different CAS levels. We first calculate the CAS for all samples and rank them. Samples with higher CAS correspond to rarer visual features, thus posing greater challenges to the model's discriminative ability. Based on the CAS value of each sample, we divide the test set into three subsets:

\begin{itemize}
    \item \textbf{High subset}: Contains the top $40\%$ of samples with the highest CAS values, representing the most challenging samples due to their rare visual features.
    \item \textbf{Middle subset}: Includes the next $30\%$ of samples, with moderate CAS values, representing samples with less rare visual features compared to the first subset.
    \item \textbf{Low subset}: Consists of the remaining $30\%$ of samples with the lowest CAS values, representing the easiest samples with more common visual features.
\end{itemize}

For each subset, we calculate and test the classification accuracy of the model before and after the improvements. This sparsity-based subset division allows us to more precisely analyze the model's performance under varying information conditions, particularly in terms of classification ability between low sparsity (information-rich) and high sparsity (information-scarce) samples, as well as the differences in model enhancement. Through this method, we can effectively assess the model's robustness and generalization when faced with samples of varying information density.

\subsection{Selection of Hyperparameter $b$}
\label{sec5.4}

\begin{figure}[t]
\begin{center}
\centerline{\includegraphics[width=1\columnwidth]{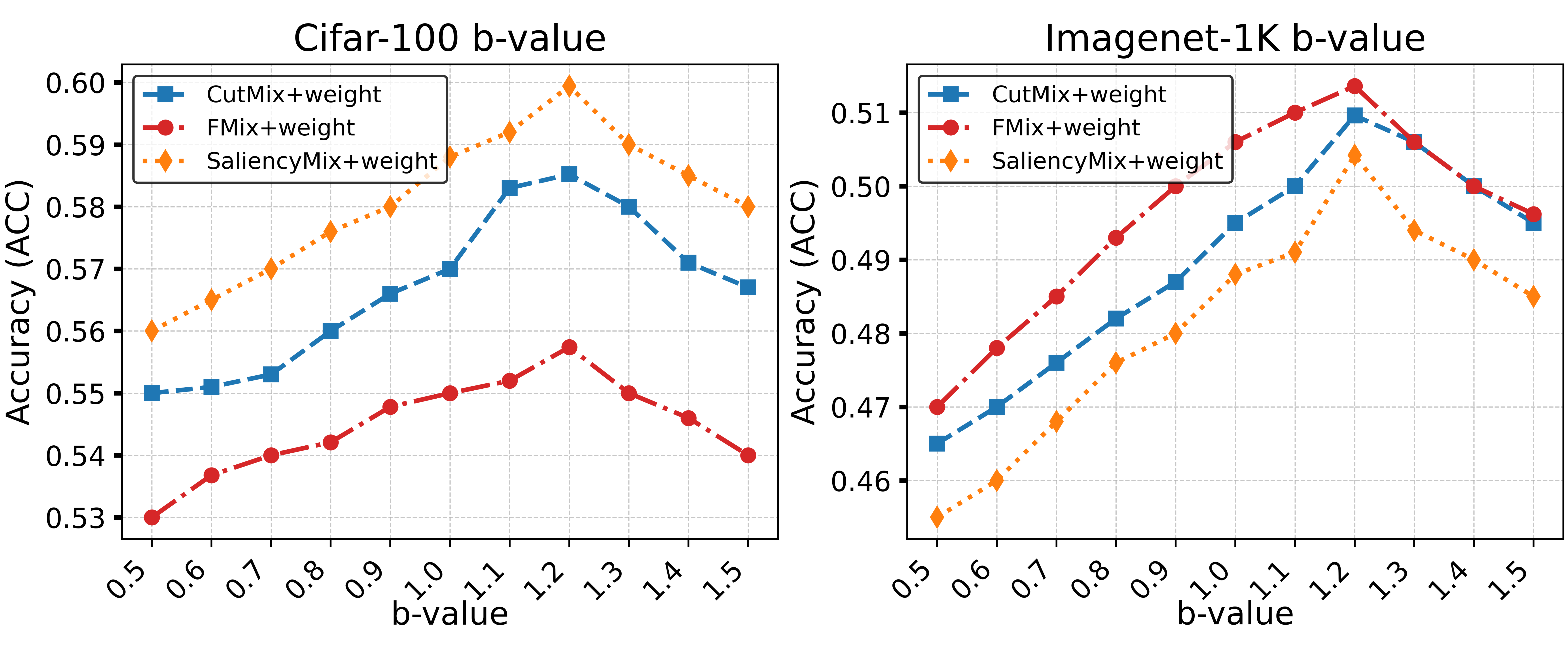}}
\vskip -0.15in
\caption{Performance of ResNeXt-50 with our method combined with CutMix, Fmix, and SaliencyMix under different values of $b$.}
\label{fig5}
\end{center}
\vskip -0.3in
\end{figure}

The power parameter \( b \) of scarcity augmentation controls the nonlinear amplification of the compositional attribute scarcity. We explored the optimal value of \( b \) by setting it within the range of 0.5 to 1.5 on \texttt{CIFAR-100} and \texttt{ImageNet}. As shown in Figure \ref{fig5}, when \( b = 1.2 \), our method achieves the highest performance gains for CutMix, FMix, and SaliencyMix. Therefore, we set \( b = 1.2 \) for all subsequent experiments.

\subsection{Main Results}
\label{sec5.5}

Table \ref{tab2} shows the classification results of the model before and after improvements across different datasets. We observed that on all datasets, the performance improved after applying our method, demonstrating the effectiveness of our approach in mitigating attribute imbalance. Impressively, with just our sampling strategy, on \texttt{ImageNet-1k} using \texttt{ResNeXt-50} as the backbone network, our method improved the overall performance of CutMix, FMix, and SaliencyMix by 1.18\%, 1.58\%, and 3.07\%, respectively. This highlights the necessity of addressing the combined attribute imbalance issue in general-purpose vision datasets.  

In fine-grained classification tasks (e.g., \texttt{Stanford Dogs, Stanford Cars, and Oxford-102 Flower}), the performance improvement was most pronounced. Specifically, on \texttt{Stanford Dogs}, using ResNeXt-50 as the backbone network, our method improved the overall performance of CutMix, FMix, and SaliencyMix by $5.43\%$, $4.83\%$, and $1.85\%$, respectively. On \texttt{Stanford Cars}, using \texttt{ResNet-18} as the backbone network, our method achieved performance gains of $7.08\%$, $10.79\%$, and $6.54\%$ for CutMix, FMix, and SaliencyMix, respectively. On \texttt{Oxford-102 Flower}, using \texttt{ResNet-18} as the backbone network, our method improved the overall performance of CutMix, FMix, and SaliencyMix by $4.99\%$, $3.21\%$, and $2.01\%$, respectively.

\subsection{Impact on Rare Samples}
\label{sec5.6}

To further analyze the effectiveness of our method across different sparsity levels, we divided the test set into three subsets: high sparsity, medium sparsity, and low sparsity, and evaluated the classification accuracy for each subset. As shown in Table \ref{tab2}, standard data augmentation methods perform poorly on high-sparsity samples, leading to a significant performance gap between low-sparsity and high-sparsity samples. However, after applying our sparsity-based sampling strategy, we observed a notable improvement in classification accuracy for high-sparsity samples, effectively reducing the performance gap between low- and high-sparsity samples.  

For instance, on \texttt{ImageNet-1k}, using \texttt{ResNeXt-50} as the backbone network, our method improved the performance of CutMix, FMix, and SaliencyMix on the high-sparsity subset by $3.54\%$, $2.61\%$, and $4.89\%$, respectively. On the fine-grained image dataset \texttt{Stanford Dogs}, with ResNet-18 as the backbone, our method enhanced the performance of CutMix, FMix, and SaliencyMix on the high-sparsity subset by $3.65\%$, $4.02\%$, and $6.64\%$, respectively. Similarly, on \texttt{Stanford Cars}, our method boosted the performance of CutMix, FMix, and SaliencyMix on the high-sparsity subset by $6.66\%$, $6.89\%$, and $6.88\%$, respectively. These results demonstrate that sparsity-guided data augmentation effectively improves the model’s ability to represent sparse attributes.  
In summary, our experimental results validate the effectiveness of the sparsity-guided data augmentation approach across multiple datasets and augmentation techniques. This method not only enhances overall classification performance but also significantly improves the model’s performance on sparse attribute samples, providing an effective solution to address attribute imbalance issues in real-world applications.

\section{Conclusion}

In this work, we explore the impact of visual attribute imbalance on image classification and propose a sampling strategy based on compositional attribute scarcity (CAS) to improve model performance on rare attributes. By integrating CAS-based sampling with data augmentation techniques, our method effectively enhances the representation of underrepresented attributes. Extensive experiments on twelve benchmark datasets validate its effectiveness in improving both robustness and fairness. Our findings emphasize the importance of attribute-aware learning and provide insights for future research on long-tailed and imbalanced learning.

\section*{Impact Statement}
This paper presents work whose goal is to advance the field of 
Machine Learning. There are many potential societal consequences 
of our work, none which we feel must be specifically highlighted here.

\nocite{langley00}

\bibliography{example_paper}
\bibliographystyle{icml2025}

\end{document}